\documentclass[3p]{elsarticle}

\usepackage{lineno,hyperref}
\modulolinenumbers[5]

\usepackage{tabularx}
\usepackage{lineno}

 \usepackage{float}
\newenvironment{applyredcolor}{\par\color{black}}{\par} % apply color to algorithm
\usepackage{xcolor}
\usepackage{textcmds}
\usepackage[linesnumbered,ruled,vlined]{algorithm2e}
\SetCommentSty{mycommfont}
\SetKwInput{KwInput}{Input}                % Set the Input
\SetKwInput{KwOutput}{Output}              % set the Output
\SetKwInput{KwParameters}{Parameters}              % set the Parameters

\journal{Journal of \LaTeX\ Templates}

\begin{document}

\begin{frontmatter}
	
	\title{RweetMiner: Automatic Identification and Categorization of Help Requests on Twitter during Disasters}

	\author[1]{Irfan Ullah}
	\ead{irfan@khu.ac.kr}

	\author[2]{Sharifullah Khan}
	\ead{sharifullah.khan@seecs.edu.pk}

	\author[3]{Muhammad Imran}
	\ead{mimran@hbku.edu.qa}

\author[1]{Young-Koo Lee\corref{cor1}%
	}
\ead{yklee@khu.ac.kr}

\cortext[cor1]{Corresponding author}

\address[1]{Department of Computer Science and Engineering, Kyung Hee University (Global Campus), Yongin 1732, South Korea}

\address[2]{School of Electrical Engineering and Computer Science (SEECS), National University of Sciences and Technology (NUST), Islamabad, Pakistan}

\address[3]{Qatar Computing Research Institute (QCRI), Doha, Qatar}

\newpageafter{author}

\begin{abstract}
Catastrophic events create uncertain environments in which it becomes very difficult to locate affected people and provide aids. People turn to Twitter during disasters for requesting help and/or providing relief to others than their friends and family. A huge number of posts issued online for seeking help could not properly be detected and remained concealed because tweets are noisy and stinky. Existing systems lack in planning an effective strategy for tweet preprocessing and grasping the contexts of tweets. This research first of all formally define request tweet in the context of social networking sites, so-called rweets, along with its different primary types and sub-types. Then the work delves into tweets for identification and categorization of rweets. For rweet identification, the precision of 99.7\% achieved using the rule-based approach and F1-measure of 82.38\% achieved using logistic regression. Logistic regression also outperformed by gaining an excellent F1-measure of 94.95\% in rweet categorization by classifying rweets into medical, volunteer, cloth, food, shelter, and money. Compared to the previous studies, a significant performance improvement is achieved for both identification and classification of rweets. We also introduced an architecture to store intermediate data to accelerate the machine learning classifiers' development process.
\end{abstract}


\end{frontmatter}

\section{Introduction}
The popularity of social networking sites (SNS) such as Facebook and Twitter has rapidly increased in recent years. %Mining on social media has been attracting many researchers due to its popularity. 
SNS have been considered as a vital source of low-latency data and thus attracted many researchers to explore their applications in different domains ranging from health informatics to sentiment and opinion mining to event detection~\cite{paul2011you,pak2010twitter,weng2011event}, to name a few. The applications of SNS for disaster response and management tasks have been well acknowledged~\cite{castillo2016big,imran2015processing}. Many research works show that SNS, in particular Twitter, contains various types of information useful for response organizations. Such information includes reports of injured or dead people, infrastructure damage, requests of needs, and donation offers~\cite{purohit2013emergency,imran2013extracting,purohit2014identifying}. 
%Due to the popularity of SNS and rapid availability of its data, researchers have explored its applications to different domains ranging from health informatics (e.g., detecting the spread of Ebola) to sentiment analysis (e.g., measuring peoples sentiment), and event detection (e.g., protests).   
%
%researchers have studied different aspects of social media mining, such as question identification \cite{dent2011through,ozger2014question,li2012question,hasanain2014identification}, and generating a taxonomy of questions \cite{gazan2011social}. The popularity of social media in disasters has also attracted the researchers for information seeking   \cite{purohit2013emergency,purohit2014identifying}. 
%
Furthermore, during disasters and emergencies, people located in or near the disaster area use social media platforms to post situational updates~\cite{cameron2012emergency}. These updates include requests for urgent needs such as food, water, shelter, etc. of affected individuals~\cite{joshi2015atwitter,goel2015asparis}. In Table \ref{tab:Request_tweets}, we show a few Twitter messages posted during Japan's Tsunami and the USA's Hurricane Sandy disasters. These messages clearly show different types of requests for help. People ask for different types of reliefs that are directly related to either their own or the lives and health of their loved ones. Some requests are very critical, i.e., asking for blood, food, or release from being trapped in dangerous situations, and should be handled immediately on an urgent basis. This data can be very useful for both people and organizations that provide relief during catastrophic events. Research studies have shown that timely access to such useful information can enable humanitarian organizations to plan relief efforts and help disaster victims. For instance, rapid identification of urgent needs of affected people and other types of requests can lead to better resource and aid planning for decision-makers. These mined tweets can potentially benefit responders facing difficulties in resources' distribution during crises \cite{zeimpekis2014human}. 

\begin{table}
	\centering
	\caption{\label{tab:Request_tweets}Request Tweets Observed during Disasters}
	\begin{tabularx}{340pt}{lX} \hline 
		S\# & Tweets Requesting for Help \\ \hline 
		1 & \qq{We're on the 7th floor of Inawashiro Hospital, but because of the risen sea level, we're stuck. Help us!!} \\ 
		2 & \qq{Help my younger brother. He called me that he is under a 4  broken house and since I live in a remote place, I can't go there. His address is (including building/apt. number).} \\ 
		3 & \qq{Thirty people are stuck at Ozaki shrine. It seems the roads are shut down. Anybody, please call -police and fire department. Anyways, I'm OK.} \\ 
		
		4 & \qq{So many in need of food supply. Help hurricane sandy victims http://t.co/1gynchqy  http://t.co/1gynchqy}\\ 
		
		5 & \qq{Stuck at east coast \#hurricanesandy. Home destroyed. Need shelter} \\ 
		
		6 & \qq{hear blood donation request red cross , shortage due hurricane sandy . please get donate.} \\

		7 & \qq{people lose everything hurricane sandy . please go donate clothes , blanket , food , etc . it 's good cause !} \\ \hline 
	\end{tabularx}
\end{table}

Despite the fact that Twitter contains valuable information during disasters, processing and extracting actionable information from tweets (e.g., urgent requests of affected people) is a challenging task. One way is to manually inspect, analyze and filter tweets for extracting actionable tweets during mass convergence scenarios, but it is not possible due to too high volume of tweets and limited resources. %of million tweets per hour. 
This problem motivated our work and a need to build an automated system that can meet the fast pace of Twitter and mine actionable tweets without having any dependency on rigid formats. Tweets are short, i.e., the maximum allowed length for a tweet is 280 characters and many tweets do not even reach the maximum limit. %Due to the limited content of tweets, the difference between contents of tweets of different classes is not obvious. 
Moreover, people use informal and brief language in tweets which is often full of shortened words. They do not care about the correctness \cite{metcalfe2010atheself}, or follow any grammatical rules/ standard structures \cite{rosa2009TextClassif}, and usually tend towards making mistakes \cite{dent2011through,metcalfe2010atheself}. Spelling mistakes can also be frequently observed in tweets. \cite{damerau1964specllcheck} states that up to 80\% of these mistakes occur due to these causes, i.e., insertion or deletion of a new character, the substitution of one character with another, and transposition or switching of two characters. Misspelling of the first character of a word, the occurrence of strong adjacency effects of characters in the keyboard, and the incident of strong character frequency can also lead to spelling mistakes. Furthermore, tweets are not formally expressive. %as it is defined in the Cambridge \textcolor{red}{and Oxford dictionaries \cite{meanOfRequestCamb, meanOfRequestOxford}.} 
Compared to standard web documents and articles which usually contain hundreds of words, understanding the semantics of tweets is challenging. Moreover, most of the text classification and state-of-the-art natural language processing techniques that are originally developed to process large documents do not perform well when applied to tweets. The existing systems employ simple data preprocessing which does not appropriately clean the noisy data and therefore, they are not achieving appropriate performance.

Therefore, our first objective is to redefine a request, named as \qq{rweet} or \qq{request tweet}, in the context of social networking sites and crises by extending the formal definition of a request provided both in the Cambridge and Oxford dictionaries. Along with the redefinition of request, we also defined three primary types and two sub-types of a request. To the best of our knowledge, this is the first study that defines this concept along with its primary types and sub-types. To deal with tweets noise, an effective data preprocessing is required to train efficient classifiers 
%for training because the use of noisy and garbage training data lead the classifiers to generate unsatisfactory results 
\cite{uysal2014impact,doct2018GIGO,Bren2018GIGO}. Therefore, the second objective of this work is to plan an efficient strategy of tweets preprocessing for producing better results. In addition, we present an extensive evaluation of the operations involved in data preprocessing. The importance and effect of every single operation in a data preprocessing is thoroughly examined from different aspects (i.e., processing time, quality of removing noisy data, the influence of one operation on another, and order of these operations) and a better data preprocessing strategy comprising an optimal number of operations along with their effective execution order is proposed to clean the data well. To handle short and concise tweets, our third objective is to grasp contexts of short texts of tweets by considering deep features to effectively distinguish among look-alike tweets that can belong to multiple classes. 
%Machine learning life cycle (MLLC) is an iterative and hit and trial process that begins with data collection and ends with the complete model that provides satisfactory performance. Preprocessed data and feature matrices are generated through computationally expensive processes that are used iteratively in the development of machine learning classifiers. 

Our fourth objective is to put forward an architecture that is effective and efficient in the development of disaster management systems. It suggests storing intermediate data in various places for orchestration and reusability in order to accelerate the development process of the system. The intermediate data will be used in both the identification and categorization of requests. The fifth objective is to achieve excellent and fair classification ability for both identification and categorization of requests. Additionally, we have achieved significant improvements over baselines in both the classification tasks. In order to obtain all these objectives, in this work we proposed \qq{RweetMiner\footnote{source code is available at: \url{https://bit.ly/2LQ3neI}}}---R stands for \textit{Requests}, a system to mine tweets containing emergency requests for resources, services, or asking for information. The proposed system first filters out tweets that contain any type of request, and then it determines the request type of a tweet, e.g., food request, medical request, shelter request, cloth request, money request, and volunteer request. %\textcolor{red}{Along with the machine learning approach, a traditional rule-based approach is also developed and tested.} 
The system has been thoroughly evaluated using disaster-related datasets, and the evaluation results show the effectiveness of the proposed system and significant performance improvements over the baseline approaches.

\par The remaining paper is organized as follows: Section \ref{sec:relatedWork} describes the related work. Section \ref{sec:propopsedArchitecture} describes the proposed architecture. Section \ref{sec:experimentsAndEvaluation} contains the experiments and evaluation. The paper has been concluded along with the future work in section \ref{sec:conclusion}.

\section{Related work} \label{sec:relatedWork}
Identification and categorization on Twitter are not novel, and studied vastly, not for the underlying problem but for an analogous domain of question mining \cite{dent2011through,ozger2014question,li2012question,hasanain2014identification, efron2010questions,gazan2011social}.
Therefore different approaches in the question mining domain are first explored and then request mining systems are reviewed in this section.

In \cite{dent2011through}, authors considered question identification as a research problem and developed a specific pipeline consists of three tools, i.e., a tokenizer, a customized lexicon, and a parser to handle complexities of the language used on Twitter, and to detect tweets with questions. The parser was implemented using 500 context-free rewrite rules to detect questions. In a rule-based approach different set of rules, for example, a rule, i.e., \qq{tweet starting with 5W1H words} is used to detect question tweets, and if a tweet satisfies one or more rules then it was considered as question tweets \cite{efron2010questions,ozger2014question,li2012question,purohit2014identifying}. Along with the rule-based approach, machine learning techniques have also been utilized for question identification. In \cite{li2012question} authors adopted two-stage cascade process for question identification. In the first stage, both traditional rule-based and machine learning techniques have been used to detect interrogative tweets (i.e., tweets containing any types of questions). But, all of these questions were not asking for seeking information, but some questions were providing information, too. For instance, \textit{\qq{Want 2 kill my boredom! Checked my mobile and the result? Insanity! Low battery.} }look like a question but it did not ask for information but provides information. Therefore, a new term \qq{qweet} was introduced for those interrogative tweets which really solicit information or request for help, and extracted in the second stage of the process. In order to mine qweets, tweet-specific features like retweets as well as context-specific features like short URLs were used as features using random forest and 10 fold cross-validation. In \cite{ozger2014question} authors adopted the same approach by developing four groups of very flexible rules, e.g., question mark and question affixes to detect maximum questions as candidate tweets. A conditional random field (CRF) with 73 features was, then used to classify tweets as \qq{question}\ and \qq{not a question} tweets. Work on qweet identification is extended in \cite{hasanain2014identification} by using 6 different types of features, i.e., tweet-specific, structural, formality, question specific, lexical, and question phrases to train an SVM classifier. In lexical features, unigrams and bigrams were used individually while question phrases features comprise of question phrases that were extracted from tweets to use as features.

In a study, \cite{purohit2013emergency} request and offers were extracted and then matched with each other in order to facilitate help and relief efforts. To clean the dataset non-ASCII characters, and stop words were removed. stemming was used to normalize the text, while tweets tags (e.g., mentions) were normalized by replacing them with specific keywords (e.g., \_MEN\_). Random forest classifier with uni-, bi-, and tri-grams individually and with additional binary features was tested for classification. Along with n-grams, contextual features (i.e., mentions, URLs, location of the person who tweeted and hashtags) and 20 topics using LDA (Latent Dirichlet allocation) were generated for each tweet to use them as features \cite{nazer2016finding}. Punctuations, stop words, duplicate tweets, retweets, and tweets having just one-word difference were removed in order to clean the dataset while classifiers, i.e., SVM, decision tree, random forest, and AdaBoost were used for performing classification.  A system ``Artificial Intelligence for Disaster Response (AIDR)'' \cite{imran2014aidr} has been implemented to take out tweets from Twitter and automatically categorized them into user-defined categories. Both humans and machines have to work together to perform the desired classification task. It comprised of three parts named collector, tagger, and trainer. The first component collects the tweets from Twitter which are then categorized by the tagger into user-defined categories. People are requested to label the subset of tweets in order to prepare training instances. These instances are then used as a training data for the classifier's training by utilizing uni- and bi-grams features individually in order to make it able for performing the desired task in real-time.

In order to extract desired information from Twitter, data should be cleaned because data on Twitter is very stinky and full of noise. Using garbage data in the training badly affects the classifier's performance \cite{doct2018GIGO,Bren2018GIGO}. Therefore, the data should be cleaned and purified well by adopting suitable preprocessing operations. Although inspiring studies \cite{purohit2013emergency, nazer2016finding} used different operations in data preprocessing, but still some noise remains in the data. Tweets posted in a language other than the focus of a study were not removed. As Twitter supports 34 different languages \footnote{https://bit.ly/2NiPDoM} therefore, all tweets written other than the language of the focus of study should be removed \cite{laylavi2017event, ozger2014question}. These tweets not only increase the size of the feature vector but also hurts the performance. Studies in \cite{purohit2013emergency}, and \cite{nazer2016finding} also failed to normalize tokens/words having  difference in only capital/lower cases (i.e., \qq{Twitter}, \qq{TWITTER}, and \qq{twitter}). This also leads to the curse of dimensionality regardless of any language, and decreases the performance, therefore should be eliminated \cite{uysal2014impact}. Morphology and syntactic variation of words should be normalized in an efficient way in order to reduce the curse of dimensionality and increasing performance. Authors in \cite{nazer2016finding} did not handled the morphology, while in \cite{purohit2013emergency} stemming has been used, but studies in \cite{korenius2004stemVsLemma}, and \cite{samir2018stemLemma} show that lemmatization provides good performance as compared to the stemming. After preprocessing, some tweets become too short to contain sufficient information (e.g., zero or single word length). These decrease the performance by generating weak features. Tweets having zero word length \cite{ozger2014question}, or a single word length should be removed. These insufficient content tweets were not removed by both \cite{purohit2013emergency} and \cite{nazer2016finding}. Some studies, i.e., \cite{laylavi2017event}, and \cite{ozger2014question} removed the punctuation while \cite{purohit2013emergency} failed to omit punctuations which unnecessarily increases the features vector space. Retweets should be handled carefully because people frequently retweet request tweets seeking help for themselves or others during disastrous scenarios \cite{starbird2010pass}. Being its too importance, \cite{nazer2016finding} removed retweets. Contamination of data duplication harms the performance and was not handled by \cite{purohit2013emergency}. Along with different operations in data preprocessing for text cleansing, the order of these operations may also affect the data cleansing operation. For using n-grams as features, there are many methods to generate them\cite{n-grams}. Although \cite{purohit2013emergency}, and \cite{nazer2016finding} adopted n-gram approach, but the method used to generate them is not clear.

\begin{figure}
	\centering
	\includegraphics[width=0.65\textwidth]{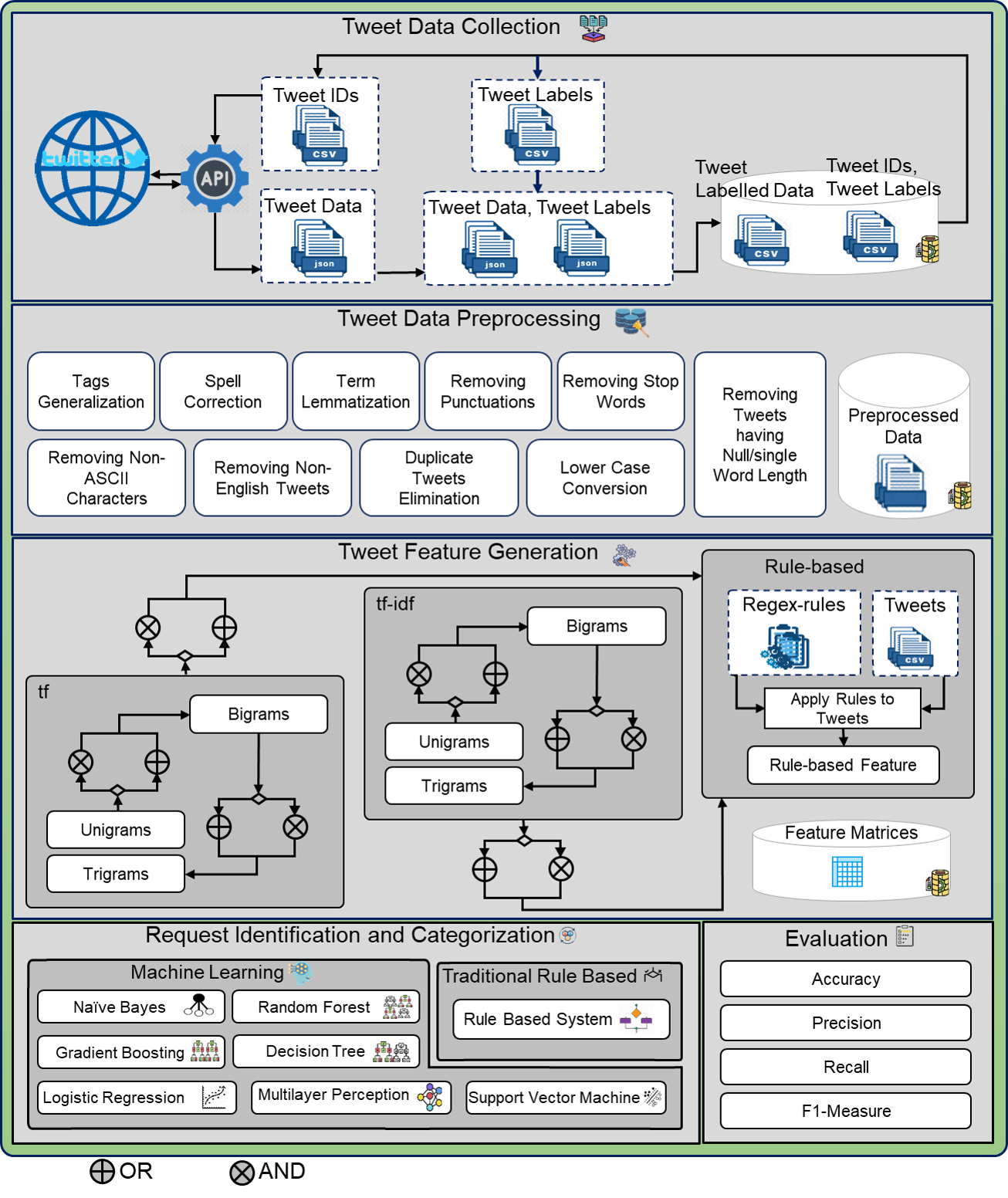}
	\caption{Architecture of the RweetMiner}
	\label{fig:method_overview}
\end{figure}
\section{Proposed design and architecture of RweetMiner}\label{sec:propopsedArchitecture}
The proposed architecture consists of four major components, as shown in Figure \ref{fig:method_overview}. The first component data collection collects the tweets by interacting with Twitter API for extracting data. The second component cleans the data in order to remove its impurities. The feature generation component extracts and generates features in the data. Finally, the last component first identifies request-related tweets and then categorizes them into specific types of requests. To describe these components with details in the following subsections, rweet has been properly defined in the first place.

\begin{table}[H]
		\caption{\label{tab:Rweet_types}Different Types of Rweets in RweetMiner}
	\begin{tabularx}{340pt}{llX} \hline 
		Type & Sub-Type & Description \\ \hline
		Declarative & - & I have an injury, need first aid box \\ 
		Interrogative & - & Can/could/may/might you bring first aid box? \\ 
		Imperative & Command & Get me first aid box \\ 
		& Request & Bring me first aid box, please \\ \hline 
	\end{tabularx}
\end{table}

\subsection{Rweet description}
According to both Cambridge and Oxford dictionaries, a request means \qq{asking for something or someone to do something, and done in a polite or official way\cite{meanOfRequestOxford, meanOfRequestCamb}.} However, people use highly creative and irregular language while authoring tweets, and express them in a very unstructured, informal, and concise way. They do not care about the correctness \cite{metcalfe2010atheself, clark2011text}, or follow any grammatical rules/ standard structures \cite{rosa2009TextClassif}, and usually tend towards making mistakes \cite{dent2011through,metcalfe2010atheself}. The possible reasons for these mistakes while authoring tweets could also be the limited length of a tweet and English as a non-native language of users \cite{damerau1964specllcheck}. In order to accommodate every possible tweet containing any type of request considering the sensitivity of disastrous situations, we cannot stick to the formal definition of request. Consequently, we examined the tweets and propose to extend the definition of a request, in the context of social networking sites (SNS), to any call expressed in any way without giving attention to politeness, formalities, and correctness. So, we name it \qq{rweet}. Rweet is a statement made in tweets that shows the deficiencies of resource(s), service(s), asking for information, or pleading for any help. It is worth noting that every request is a rweet, but a rweet may not be a proper request following formal definitions posted by Cambridge and Oxford dictionaries\cite{meanOfRequestOxford, meanOfRequestCamb}. We not only redefined the request but also defined the types and sub-types of requests made by the people. Table \ref{tab:Rweet_types} shows different possible types of expressing rweets that are considered as requests in this research. 
\subsection{Data collection}
We have used two datasets of tweets in this research that have also been used by \cite{purohit2013emergency}. Due to the terms of service of Twitter, \cite{purohit2013emergency} provided only tweet IDs along with class labels.  The accessibility of Twitter API allows fetching publically available tweets based on different criteria, e.g., tweet IDs. A python-based library Tweepy, successfully used and tested in different studies such as \cite{almatrafi2015application}, is used to fetch these tweets using tweet IDs. One of the two dataset, i.e., dataset-1 comprises 2,940 labeled tweets categorized into two classes, i.e., ``request'' (1644, 56\%) and ``not request'' (1296, 44\%). The dataset-2 comprises 2,707 tweets categorized into six classes representing different types of requests, i.e., ``money'' (1896, 70\%), ``volunteer'' (216, 8\%), ``cloth'' (165, 6\%), ``shelter'' (146, 5\%), ``medical'' (144, 5\%), and ``food'' (140, 5\%). Data collection component in Figure \ref{fig:method_overview} shows the procedure for extracting data from Twitter server.

\begin{algorithm}
	
	\DontPrintSemicolon
	\label{algo:datapreprocessing}
	\begin{applyredcolor}
		\KwInput{Actual tweets data obtained from Twitter Server }
		\KwOutput{Cleaned tweets}
		\KwParameters{$TW$: Actual tweets data, $TW_{clean}$: Preprocessed tweets, $tw_{NoASCII}$: Remove non-ASCII characters,
			$tw_{Eng}$: Remove non-English tweets, $tw_{LC}$: Lower case conversion, $tw_{NoP}$: Remove punctuations, $tw_{NoSW}$: Remove stop words, $tw_{L>1}$: Remove tweets with null or single word length, $tw_{GoT}$: Generalization of tags, $tw_{SC}$: Spell correction, $tw_{TL}$: Term lemmatization        }
		\ForEach{tweet tw in $TW$}
		{
			$tw_{NoASCII} \leftarrow$  RemovingNonASCIICharacters($tw$) \; \label{algo:datapreprocessing:RemovingNonASCIICharacters}
			$tw_{Eng} \leftarrow$ RemovingNonEnglishTweets($tw_{NoASCII}$) \; \label{algo:datapreprocessing:RemovingNonEnglishTweets}
			$tw_{LC} \leftarrow$ LowerCaseConversion($tw_{Eng}$) \;
			\label{algo:datapreprocessing:LowerCaseConversion}
			$tw_{NoP} \leftarrow$ RemovingPunctuations($tw_{LC}$) \;
			\label{algo:datapreprocessing:RemovingPunctuation}
			$tw_{NoSW} \leftarrow$ RemovingStopWords($tw_{NoP}$) \;
			\label{algo:datapreprocessing:RemovingStopWords}
			$tw_{L>1} \leftarrow$ RemovingTweetsHavingNullSinglWordLength($tw_{NoSW}$) \;
			\label{algo:datapreprocessing:RemovingTweetsHavingNullSinglWordLength}
			$tw_{GoT} \leftarrow$ \texttt{GeneralizationOfTags($tw_{L>1}$)} \;
			\label{algo:datapreprocessing:GeneralizationOfTags}
			$tw_{SC} \leftarrow$ \texttt{SpellCorrection($tw_{GoT}$)} \;
				\label{algo:datapreprocessing:SpellCorrection}
			$tw_{TL} \leftarrow$ \texttt{TermLemmatization($tw_{SC}$)} \; \label{algo:datapreprocessing:TermLemmatization}
		}
		$TW_{clean} \leftarrow$ EliminatingDuplicateTweets($TW$) \; \label{algo:datapreprocessing:EliminatingDuplicateTweets}
		Save($TW_{clean}$) \; \label{algo:datapreprocessing:Save}
		
		\SetKwFunction{FGeneralizationOfTags}{GeneralizationOfTags} 
		\SetKwProg{Fn}{}{:}{}
		\Fn{\FGeneralizationOfTags{$tw_{L>1}$}}{  \label{algo:datapreprocessing:FGeneralizationOfTags}
			$tw_{GoT} \leftarrow \emptyset$ \;
			$tw_{num} \leftarrow$ replace(r\qq{?:(?:{\textbackslash}d+,?)+(?:{\textbackslash}.?{\textbackslash}d+)?)},\qq{\_NUM\_}, $tw_{L>1}$) \;  \tcp*{Replace text matched by \q{r....} with \_NUM\_ in $tw_{num}$} \label{algo:datapreprocessing:FGeneralizationOfTags:replaceNUM}
			$tw_{rt} \leftarrow$ replace(r\qq{?:(RT{\textbar}rt) @ ?[{\textbackslash}w\_]+:?)},\qq{\_RT\_},$tw_{num}$) \;
			$tw_{mention} \leftarrow$ replace(r\qq{?:@ ?[{\textbackslash}w\_]+)},\qq{\_MENT\_}, $tw_{rt}$) \;
			$tw_{GoT} \leftarrow$ replace(r\qq{http[s]? ?: ?//(?:[a-z]{\textbar}[0-9]{\textbar}[\$-\_@.\&amp;+]{\textbar}[!*{\textbackslash}({\textbackslash}),]{\textbar}(?:\%[0-9a-f][0-9a-f]))+},\qq{\_URL\_}, $tw_{mention}$) \; \label{algo:datapreprocessing:FGeneralizationOfTags:replaceURL}
			\KwRet 	$tw_{GoT}$\;
			
		}
		
		\SetKwFunction{FSpellCorrection}{SpellCorrection}
		\SetKwProg{Fn}{}{:}{}
		\Fn{\FSpellCorrection{$tw_{GoT}$}}{ \label{algo:datapreprocessing:FSpellCorrection}
			$tw_{SC} \leftarrow \emptyset$ \;
			\ForEach{word in $tw_{GoT}$}
			{
				\If{word is not in [\qq{\_NUM\_}, \qq{\_RT\_}, \qq{\_MENT\_}, \qq{\_URL\_}]}
				{ \label{algo:datapreprocessing:FSpellCorrectionSkipTags}
					\If{ErrorDetection($word$)}
					{ \label{algo:datapreprocessing:FSpellCorrectionErrorDetection}
						$tw_{SC} \leftarrow$ replace($word$, ErrorCorrection($word$), $tw_{GoT}$) \;  \label{algo:datapreprocessing:FSpellCorrectionErrorCorrection}
						\tcp*{Replace misspelled $word$ with the corrected one in $tw_{GoT}$  }
					}
				}
			}
			\KwRet 	$tw_{SC}$\;
			
		}
		
		\SetKwFunction{FTermLemmatization}{TermLemmatization}
		\SetKwProg{Fn}{}{:}{}
		\Fn{\FTermLemmatization{$tw_{SC}$}}{  
			$tw_{TL} \leftarrow \emptyset$ \;
			\For{word in $tw_{SC}$}
			{ 
				\If{word is not in [\qq{\_NUM\_}, \qq{\_RT\_}, \qq{\_MENT\_}, \qq{\_URL\_}]} 
				{ \label{algo:datapreprocessing:FTermLemmatizationSkipTags}
					$tw_{TL} \leftarrow$ replace($word$, Lemmatize($word$), $tw_{SC
					}$)
				}
			}
			
			\KwRet 	$tw_{TL}$\;
			
		}
		
		\caption{Tweet Data Preprocessing}
	\end{applyredcolor}
\end{algorithm}

\subsection{Data preprocessing}
Data preprocessing is a very critical and core step of natural language processing problems. Tweets are full of redundant and garbage data. The use of noisy and garbage training data leads the classifiers to generate unsatisfactory results \cite{doct2018GIGO,Bren2018GIGO}, so to improve the performance, data preprocessing is vital. Therefore, an effective data cleaning strategy is proposed to generate the well preprocessed, cleaned, and efficient data to be fed to the classifiers for training. As the proposed strategy purify well the data by pruning all the noisy and dirty data which in turn reduces the memory requirements for the preprocessed data. Algorithm \ref{algo:datapreprocessing}, presents the pseudo-code for the proposed data preprocessing.
\par Not only the number of operations but also the order of these operations are also very important, and greatly affect the performance of the classifiers. Therefore, an optimized number of operations and efficient execution order of these operations have also been proposed that reduces the processing time of the data preprocessing.

To standardized the text, non-ASCII characters should be detected and removed at the very start of the data preprocessing operations (line \ref{algo:datapreprocessing:RemovingNonASCIICharacters} in algorithm \ref{algo:datapreprocessing}), because it is observed that they flow through different operations, e.g., lemmatization, stemming, spell correction, generating dependency tree, and parts-of-speech tagging, etc., and badly affect their performance. As the current study focuses only on the English language, therefore, all non-English tweets were removed (line \ref{algo:datapreprocessing:RemovingNonEnglishTweets} in algorithm \ref{algo:datapreprocessing}) because Twitter supports 34 different languages\footnote{https://bit.ly/2NiPDoM} for authoring tweets. These tweets should be removed before the language-dependent operations, i.e., lemmatization, stemming, spell correction, and stop words removal because they unnecessarily increase the processing time of the data preprocessing and memory requirements of the preprocessed data.
It was also perceived that non-English tweets also lead to the curse of dimensionality which in turn increases the training time, and features generated from non-English tweets also affect the classifiers' predictions on English tweets in a bad way when used during training. All the text was converted to lower case letters(line \ref{algo:datapreprocessing:LowerCaseConversion} in algorithm \ref{algo:datapreprocessing}). Lower case conversion proved to be an effective operation for having its positive effect on the performance\cite{uysal2014impact}. 
It also avoids the differentiation of words based on lower and upper case letters, e.g., “Hurricane”, “HURRICANE”, and “hurricane”. It also directly enhances the elimination of duplicate tweets operation and reduces the curse of dimensionality of feature vector space. After performing the operation of stop words removal (line \ref{algo:datapreprocessing:RemovingStopWords} in algorithm \ref{algo:datapreprocessing}), some tweets were left that have zero or single word lengths. These tweets should be removed(line \ref{algo:datapreprocessing:RemovingTweetsHavingNullSinglWordLength} in algorithm \ref{algo:datapreprocessing}) because of the absence of a sufficient amount of information for judging/predicting labels for them. It also reduces the processing time of the data preprocessing component by avoiding the last three operations to be executed on these tweets. Along with lemmatization, stemming was also tested and lemmatization was being proved to be more effective than stemming because sometimes stemming fails to convert different forms of a word to its basic form. Although lemmatization improved the performance, stemming should be used if the focus is the speed. As explained earlier that users extensively use different tags, i.e., hashtags, numbers, mentions, retweets, and hyperlinks in short texts of tweets. This information is not completely meaningless in the context of the disasters \cite{nazer2016finding,starbird2010pass}, therefore should not be removed completely. Unlike \cite{ozger2014question,hemalatha2012a}, these tags were generalized \cite{purohit2013emergency} by replacing numbers with \_NUM\_, URLs with \_URL\_, retweets (RT @user-name) with \_RT\_, and mentions (@ user-name) with \_MENT\_ (lines \ref{algo:datapreprocessing:FGeneralizationOfTags:replaceNUM}-\ref{algo:datapreprocessing:FGeneralizationOfTags:replaceURL} in algorithm \ref{algo:datapreprocessing}) for preserving somewhat information as well as reducing the dimensionality of feature vector space and processing time for data preprocessing.
% and memory requirements for preprocessed data. 
 Hashtags (\#tag) are very valuable in disastrous situations \cite{moore2014hashtag}, therefore they should neither be removed nor generalized.
\par During a large number of experiments, it was observed that both involved operations, as well as their execution order, affect the data cleansing process directly. For example, by ignoring the lower case conversion, the elimination of duplicate tweets operation will fail to remove tweets having lower/capital case letters' differences which in turn not only reduces the performance, as said earlier but also increase the dimensions of feature vector space. And if it is performed after the elimination of duplicate tweets operation, then those tweets will become duplicates.
%For example, ignoring lower case conversion, or performing it after elimination of duplicate tweets operation will be left the data with contaminated tweets having just lower/capital case letters difference which in turn not only reduce the performance, as said earlier but will also increase the dimensions of feature vector space. It also slightly increases the processing time and memory requirements for generating and storing preprocessed data.
It also slightly increases the processing time for generating preprocessed data. Lemmatization involves searching for a basic form (aka lemma) of a word in a huge dictionary therefore it should be performed(line \ref{algo:datapreprocessing:TermLemmatization} in algorithm \ref{algo:datapreprocessing})  after the stop words removal operation because it eliminates the processing burden of performing lemmatization operation on stop words that will be removed eventually. But, it should be performed before the elimination of duplicate tweets operation because it kills the chances of the differences between tweets based on inflected forms of words, e.g., the same output \qq{He write a letter} will be generated for \qq{He writes a letter} and \qq{He wrote a letter} texts which could be then easily treated as duplicates. Lemmatization operation should be performed after the generalization of tags operation so that it could be skipped for generalized tags (line \ref{algo:datapreprocessing:FTermLemmatizationSkipTags} in algorithm \ref{algo:datapreprocessing}), i.e., numbers, mentions, URLs, and retweets in order to improve the processing efficiency. Duplicate tweets should be removed completely because they skew a classifier towards a class with duplicate tweets because some classifiers like SVM are very sensitive towards identical tweets. Usually, this operation is performed in the beginning stages of data preprocessing. We have observed that performing it in the initial stages does not eliminate duplication completely. For example,
it will not remove texts \qq{He is going to school @akram, www.example.com} and \qq{He goes to School @ahmed, www.example123.com}. We have decided to perform it as the last step of data preprocessing (line \ref{algo:datapreprocessing:EliminatingDuplicateTweets} in algorithm \ref{algo:datapreprocessing}) before the feature extraction so that those tweets having differences in just inflection of words, stop words, numbers, URLs, mentions, or capitalization should be removed. For example, after passing the same example texts \qq{He is going to school @akram, www.example.com} and \qq{He goes to School @ahmed, www.example123.com}, from all the above steps, then they will prune to \qq{he go school \_MENT\_ \_URL\_} and \qq{he go school \_MENT\_ \_URL\_} and thus the duplication can be easily eliminated.
Therefore, care should be taken while choosing and executing operations of data preprocessing in the data cleansing process. %Here, we propose an effective preprocessing strategy consists of 9 different operations, along with their efficient order of execution to well purify the dataset, and also reducing the time and memory complexities of data preprocessing, too.
Here, we propose an effective preprocessing strategy consists of 9 different operations, along with their efficient order of execution to well purify the dataset as well as reducing the preprocessing time of data preprocessing, too. The operations and their execution order  (lines \ref{algo:datapreprocessing:RemovingNonASCIICharacters}-\ref{algo:datapreprocessing:EliminatingDuplicateTweets} in algorithm \ref{algo:datapreprocessing}) is: 1) Removing non-ASCII characters, 2) Removing non-English tweets, 3) Lower Case Conversion, 4) Removing Punctuations, 5) Removing Stop Words, 6) Removing tweets having null/single word length, 7) Generalization of Tags, 8) Term Lemmatization, and 9) Eliminating duplicate tweets. 

It is also worth mentioning that spell correction was also tested for the underlying problem but then ignored due to two reasons: 1) it did not improve the performance, and 2) increased the processing time too much. Spell correction is a computationally expensive process that performs two searches; one in the error detection and another one in the error correction steps. The reason for no improvement of performance might be that tweets are full of different categories of errors while the reason for slow processing is that for every single word in a large tweet data at least one of the two steps, i.e., error detection(line \ref{algo:datapreprocessing:FSpellCorrectionErrorDetection} in algorithm \ref{algo:datapreprocessing})  and correction(line \ref{algo:datapreprocessing:FSpellCorrectionErrorCorrection} in algorithm \ref{algo:datapreprocessing}) of spell correction(line \ref{algo:datapreprocessing:FSpellCorrection} in algorithm \ref{algo:datapreprocessing}) should be performed.
%The reason for no improvement of performance might be that tweets are full of different categories of errors while the reason for slow processing is that for every single word, in a huge vocabulary of tweet data, spell correction operation (line \ref{algo:datapreprocessing:FSpellCorrection} in algorithm \ref{algo:datapreprocessing}) is performed in two steps; error detection(line \ref{algo:datapreprocessing:FSpellCorrectionErrorDetection} in algorithm \ref{algo:datapreprocessing})  and correction(line \ref{algo:datapreprocessing:FSpellCorrectionErrorCorrection} in algorithm \ref{algo:datapreprocessing}).
These nine types of mistakes can be observed frequently in tweets, i.e., changed letter (“monthz” for “months”), dropped letters (e.g., “runnin” for “running”), acronym (lol for “laugh out loud” ), Misspelling (e.g., “marls” for ‘marks”), punctuation error (e.g., “hes” for “he’s” ), non-dictionary slang (e.g., “this thing was well mint” (this thing was very good)), repeated letters (e.g., I am sorrrrrrrrrry (I am sorry)), homophones ( “wait” for “w8”) and emoticons (e.g.,  “:)” for (a smiling face))\cite{clark2011text}. It is also worth to mention that spell correction can be placed in two places in the operations' execution order of data preprocessing; after removing non-English tweets (line \ref{algo:datapreprocessing:RemovingNonEnglishTweets} in algorithm \ref{algo:datapreprocessing}) or before lemmatization(line \ref{algo:datapreprocessing:TermLemmatization} in algorithm \ref{algo:datapreprocessing}). If cleanliness and purity is the focus then former one should be preferred because it will improve the performance of removing stop words, term lemmatization, and tokenization at the cost of increased processing time, and later one is recommended if the focus is the speed because it will avoid applying spell correction operation on stop words, which will be removed later,  and the generalized tags in tweets (line \ref{algo:datapreprocessing:FSpellCorrectionSkipTags} in algorithm \ref{algo:datapreprocessing}). %Applying the pruning operations in the correct time and place leaves no chance for garbage data to be remain, as explained previously in the example. The proposed data preprocessing methodology not only purifies the data and reduces the time complexity of the data preprocessing but also improves the memory complexity of the preprocessed data. 
Applying the pruning operations in the correct time and place leaves no chance for garbage data to be remained, as explained previously in the example. The proposed data preprocessing approach not only improves the processing time of data preprocessing, but also purifies well the data which in turn reduces the memory requirements for the preprocessed data.

The preprocessed and purified data is then written to the disk for re-usability and orchestration (line \ref{algo:datapreprocessing:Save} in algorithm \ref{algo:datapreprocessing} and line \ref{algo:featureextraction:Load} in algorithm \ref{algo:featureextraction}), as shown in the data preprocessing component in Figure \ref{fig:method_overview}. Machine learning life cycle (MLLC) is an iterative and hit and trial process in which preprocessed data is used extensively for generating different types of feature matrices for training, tuning, and evaluating classifiers' performance. Therefore the persistence preprocessed data could be used repeatedly for feature generation in the development of effective single or multiple machine learning classifiers to avoid repetitive re-computation of the computationally expensive process of data processing.

\begin{algorithm}
\small
\begin{applyredcolor}
	\DontPrintSemicolon
	\label{algo:featureextraction}
	\KwInput{Actual tweets data obtained from Twitter server and Preprocessed tweets data}
	\KwOutput{Set of feature matrices along with labels}
	\KwParameters{$TW$: Actual tweets data, $TW_{clean}$: Preprocessed tweets, $TW_f$: Sets of feature matrices along with labels,
	$SPs$: Set of sequential patterns, $TW_r$: Rule-based features, $tw_{f}$: Single feature matrix, 
min\_freq: minimum frequency for filtering terms, max\_freq: maximum frequency for filtering terms }
	%$R \leftarrow 28, C \leftarrow 2$ \;
			$TW_{clean} \leftarrow$ Load($TW_{clean}$) \; 
	\label{algo:featureextraction:Load} \tcp*{ For $TW_{clean}$ see line 12 in algorithm 1}
	$TW_f[R,C] \leftarrow \emptyset$ \;\tcp*{ R is the total number of feature matrices, and C is the total number of columns  }

	$tw_{f}[r, c] \leftarrow \emptyset$ \; \tcp*{ r = $|TW|$, and c is the total number of terms to represent each tweet }
	$i \leftarrow 1, j \leftarrow 1$ \;
	\ForEach{extendRuleFeat in [\qq{Yes}, \qq{No}]}
	{
		  \ForEach{vect in  [\qq{tf}, \qq{tf-idf}]} 
		  {\label{algo:featureextraction:tf-tf-idf}
				  	
					\For{i $\leftarrow$ 1 to 3}
					{   $tw_{f}[r, c] \leftarrow$ GenerateN-grams($TW_{clean}$, n-range $\leftarrow$ (i,i), vectorizer = vect, min\_freq $\leftarrow$ 1, max\_freq $\leftarrow$ 1) \;
				
				  		\If{extendRuleFeat == \qq{yes}}
							     {\label{algo:featureextraction:extendRuleFeatSingleNgram}
							     	$tw_{f}[r, c] \leftarrow$ Merge($tw_{f}[r, c]$, RuleBasedFeatures($TW$)) \;
							     	$TW_f[R,C]$.add(\qq{0-i-ruleFeatures}, $tw_{f}[r, c]$) \;
						     }
						     \Else
									 {
									 $TW_f[R,C]$.add(\qq{0-i}, $tw_{f}[r, c]$) \;
									 }
							 
							\For{j $\leftarrow$ i+1 to 3}
							   {\label{algo:featureextraction:combined-tf-tf-idf}
									 $tw_{f}[r, c] \leftarrow$ GenerateN-grams($TW_{clean}$, n-range $\leftarrow$ (i,j), vectorizer = vect, min\_freq $\leftarrow$ 1, max\_freq $\leftarrow$ 1) \; \tcp*{n-range =  (i,j) means that generate n-grams between the values of i and j} \label{algo:featureextraction:n-range}
									 \If{extendRuleFeat == \qq{yes}}
									 {\label{algo:featureextraction:extendRuleFeatMultiNgram}
									 $tw_{f}[r, c] \leftarrow$ Merge($tw_{f}[r, c]$, RuleBasedFeatures($TW$))\;
									 	$TW_f[R,C]$.add(\qq{i-j-ruleFeatures}, $tw_{f}[r, c]$) \;
									 }
									 \Else
									 {
									 $TW_f[R,C]$.add(\qq{i-j}, $tw_{f}[r, c]$) \;
									 }
						
					       }

		    	}
		    	
		 }
	}
	Save($TW_f[R,C]$) \; \label{algo:featureextraction:Save}

	\SetKwFunction{FRuleBasedFeatures}{RuleBasedFeatures}
	\SetKwProg{Fn}{}{:}{}
	\Fn{\FRuleBasedFeatures{$TW$}}{ \label{algo:featureextraction:FRuleBasedFeatures}
  $SPs \leftarrow$ set of all sequential patterns\;
  
	$TW_r[r, g] \leftarrow \emptyset$ \; \tcp*{ $g = |SPs|$ and $r \leftarrow |TW|$  }
	\For{i $\leftarrow$ 1 to r}
	{ 
		$j \leftarrow 1$ \;
		\ForEach{$\textrm{pattern} \textit{ p} \in SPs$}
		{
			\If{$TW[i, 1] \textrm{ satisfies} \textit{ p}$} 
			{\tcp*{ $TW[i, 1]$ are the tweets while $TW[i, 2]$ are the corresponding labels }
				$TW_r[i, j] \leftarrow 1$\;

			}
			\Else
			{
				$TW_r[i, j] \leftarrow 0$\;
			}
		 $j \leftarrow j + 1$\;
			
	  }
  }
		\KwRet 	$TW_r[r,g]$\;
		
		}

	\caption{Tweet Feature Generation}
\end{applyredcolor}
\end{algorithm}

\subsection{Feature Generation} 
In order to extract numerical features from datasets, \qq{Bag of n-grams} technique along with rule-based features have been used in this research. Algorithm \ref{algo:featureextraction} provides the pseudo-code for the feature generation. For generating n-grams, we considered n from 1 to 3. In other words, a term/feature can be uni-gram, bi-gram, and tri-gram. The use of simple n-grams is not effective in the classification, especially when the text to be classified is very short and limited like tweets. Therefore, to extract potent features from the limited texts of tweets, these n-grams were not just used in their original form, but in order to grasp the deep features, feature matrices of these n-grams have been expended by appending uni-grams with bi and/or tri-grams, and bi-grams with tri-grams (line \ref{algo:featureextraction:combined-tf-tf-idf} in algorithm \ref{algo:featureextraction}). This helps in increasing the distance among look-alike tweets (i.e, can belong to multiple classes). For example, the cosine similarity \cite{cosSim} for two sentences, i.e., \qq{He loves me} and \qq{He likes me} when computed using just uni-grams is 0.6. But using a customized extended feature matrix of n-grams, i.e., uni-grams+bi-grams, and uni+bi+tri-grams the cosine similarities drop to 0.4 and 0.333 respectively. Hence, this strategy widens the distance among look-alike tweets, and helps in accurately predicting tweets' classes; thus improves the performance. 

Term Frequency (tf) and/or Term Frequency-Inverse Document Frequency (tf-idf) have been used to generate the numerical values for the n-grams used as features (line \ref{algo:featureextraction:tf-tf-idf} in algorithm \ref{algo:featureextraction}). tf is the occurrence of each and every term in a text while tf-idf shows that a term appearing in many documents should be assigned a lower weight than a term appearing in few documents \cite{buttcher2010information}.
Frequency-based filtering are also applied to the terms in n-grams in which those terms that did not appear in a specific number of tweets or a portion of tweets data were ignored. This frequency-based filtering slightly reduced the curse of dimensionality at the cost of a small drop in performance, therefore it was ignored.

\begin{table}[H]
	\centering
	\caption{\label{tab:Rweet_patterns}List of Regular Expressions Derived From Sequential Patterns Provided by \cite{purohit2013emergency}
	}
	\begin{tabularx}{350pt}{lX} \hline 
		S\# & Patterns \\  \hline
		1 & {\textbackslash}b(I{\textbar}we){\textbackslash}b.*{\textbackslash}b(am{\textbar}are{\textbar}will be){\textbackslash}b.*{\textbackslash}b(bringing{\textbar}giving{\textbar}helping{\textbar}raising{\textbar}donating{\textbar}  auctioning){\textbackslash}b \\ 
		2 & {\textbackslash}b(I{\textbackslash}'m){\textbackslash}b.*{\textbackslash}b(bringing{\textbar}giving{\textbar}helping{\textbar}raising{\textbar}donating{\textbar} auctioning){\textbackslash}b \\  
		3 & {\textbackslash}b(we{\textbackslash}'re){\textbackslash}b.*{\textbackslash}b(bringing{\textbar}giving{\textbar}helping{\textbar}raising{\textbar}donating {\textbar}auctioning){\textbackslash}b \\  
		4 & {\textbackslash}b(I{\textbar}we){\textbackslash}b.*{\textbackslash}b(will{\textbar}would like to){\textbackslash}b.*{\textbackslash}b(bring{\textbar}give{\textbar}help{\textbar}raise{\textbar}donate{\textbar}auction){\textbackslash}b \\  
		5 & {\textbackslash}b(I{\textbar}we){\textbackslash}b.*{\textbackslash}b(will{\textbar}would like to){\textbackslash}b.*{\textbackslash}b (work{\textbar}volunteer{\textbar}assist){\textbackslash}b \\ 
		6 & {\textbackslash}b(we{\textbackslash}'ll){\textbackslash}b.*{\textbackslash}b(bring{\textbar}give{\textbar}help{\textbar}raise{\textbar}donate{\textbar}auction){\textbackslash}b \\  
		7 & {\textbackslash}b(I{\textbar}we){\textbackslash}b.*{\textbackslash}b(ready{\textbar}prepared){\textbackslash}b.*{\textbackslash}b(bring{\textbar}give{\textbar}help{\textbar}raise{\textbar}donate{\textbar}auction){\textbackslash}b \\
		8 & {\textbackslash}b(where){\textbackslash}b.*{\textbackslash}b(can I{\textbar}can we){\textbackslash}b.*{\textbackslash}b(bring{\textbar}give{\textbar}help{\textbar} raise{\textbar}donate){\textbackslash}b \\  
		9 & {\textbackslash}b(where){\textbackslash}b.*{\textbackslash}b(can I{\textbar}can we){\textbackslash}b.*{\textbackslash}b(work{\textbar}volunteer {\textbar}assist){\textbackslash}b \\ 
		10 & {\textbackslash}b(I{\textbar}we){\textbackslash}b.*{\textbackslash}b(like{\textbar}want){\textbackslash}b.*{\textbackslash}bto{\textbackslash}b.*{\textbackslash}b(bring{\textbar}give{\textbar}help{\textbar}raise{\textbar}donate){\textbackslash}b \\  
		11 & {\textbackslash}b(I{\textbar}we){\textbackslash}b.*{\textbackslash}b(like{\textbar}want){\textbackslash}b.*{\textbackslash}bto{\textbackslash}b.*{\textbackslash}b(work{\textbar}volunteer{\textbar}assist){\textbackslash}b \\  
		12 & {\textbackslash}b(will be){\textbackslash}b.*{\textbackslash}b(brought{\textbar}given{\textbar}raised{\textbar}donated{\textbar} auctioned){\textbackslash}b \\ 
		13 & {\textbackslash}b{\textbackslash}w*{\textbackslash}s*{\textbackslash}b{\textbackslash}? \\ 
		14 & {\textbackslash}b(you{\textbar}u).*(can{\textbar}could{\textbar}should{\textbar}want to){\textbackslash}b \\
		15 & {\textbackslash}b(can{\textbar}could{\textbar}should).*(you{\textbar}u){\textbackslash}b \\ 
		16 & {\textbackslash}b(like{\textbar}want){\textbackslash}b.*{\textbackslash}bto{\textbackslash}b.*{\textbackslash}b(bring{\textbar}give{\textbar}help{\textbar}raise{\textbar}donate){\textbackslash}b \\ 
		17 & {\textbackslash}b(how){\textbackslash}b.*{\textbackslash}b(can I{\textbar}can we){\textbackslash}b.*{\textbackslash}b(bring{\textbar}give{\textbar}help{\textbar}raise{\textbar}donate){\textbackslash}b \\ 
		18 & {\textbackslash}b(how){\textbackslash}b.*{\textbackslash}b(can I{\textbar}can we){\textbackslash}b.*{\textbackslash}b(work{\textbar}volunteer{\textbar} assist){\textbackslash}b \\ \hline 
	\end{tabularx}
\end{table}

\par Eighteen salient sequential patterns, manually extracted by Red Cross experts by studying a large number of request tweets \cite{purohit2013emergency}, have been used for generating features. Regular expressions were developed for these patterns as shown in Table \ref{tab:Rweet_patterns}. Each of these regular expressions has been used as features in addition to n-grams (lines \ref{algo:featureextraction:extendRuleFeatSingleNgram} and \ref{algo:featureextraction:extendRuleFeatMultiNgram} in algorithm \ref{algo:featureextraction}). To do so, each and every regular expression was checked against each tweet, if it satisfies then its frequency is 1, otherwise 0 \cite{purohit2013emergency,cong2008finding}. We named these features \qq{rule-based features} (\ref{algo:featureextraction:FRuleBasedFeatures} in algorithm \ref{algo:featureextraction}). Explanation and interpretation of the special characters\cite{regExp} are provided in Table \ref*{tab:RE_CharExp}.
Finally, tf and tf-idf feature vectors have been normalized using euclidean norm (L2-norm) because it speeds up the training time and improves the classifiers' performance \cite{suarez2012statistical, sing2020invest, jin2015data}. Table \ref{tab:Features_Combinations} shows the 24 unique combinations of these features used in this research. Each feature vector is a very large and sparse matrix $M(R \times C)$ (where $R$ refers to the total number of tweets and $C$ refers to the total number of terms/tokens/features or size of vocabulary) that is given as an input along with label vector to the classifiers for training. In $M$, each tweet is represented by a $1 \times C$ vector of features.
\par
\begin{figure}[H]
	\centering
	\includegraphics[width=1\textwidth]{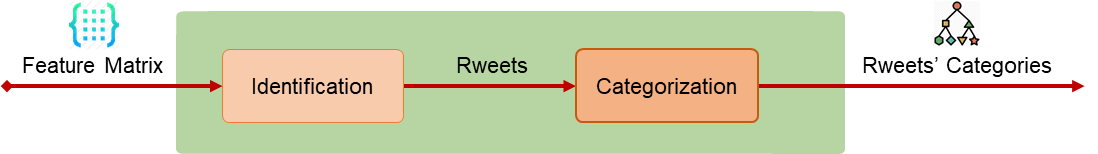}
	\caption{Rweet Identification and Rweet Categorization in a Series }
	\label{fig:classificationseries}
\end{figure}

 As MLLC is an iterative process in which the classifier is tested and tuned on different sets of feature matrices for maximizing performance by avoiding overfitting and underfitting problems. Therefore, in order to avoid the re-computation of these feature matrices, they should be stored on the disk for orchestration and re-usability (line \ref{algo:featureextraction:Save} in algorithm \ref{algo:featureextraction} and line \ref{algo:rweetseries:Load} in algorithm \ref{algo:rweetseries}), as shown in the feature generation component in Figure \ref{fig:method_overview}. These persistent feature matrices can be used iteratively in the processes of feature construction, dimensionality reduction, tuning, debugging, and evaluation of many classifiers, simultaneously. These persistent feature matrices should be be utilized and flowed through both rweet identification and rweet  categorization phases in a series, as shown in Figure \ref{fig:classificationseries}, instead of re-computing them repetitively for each phase.
\par Descriptive statistics\cite{tweetLength} show that only a small portion of 1\% of tweets touches the maximum allowed characters limited. People author tweets in a very concise way, only 5\% of tweets' length exceeds 190 characters and the most usual length for a tweet is 33 characters. As $C$ (in $1 \times C$ vector) would be very large because it represents the whole vocabulary of the dataset and tweets will contain a very small subset of tokens, the resultant matrix will be very sparse by containing a large number of zeros (more than 99\%). For example, a dataset with 1000 tweets will generate more than 20,000 different unique set of tokens in total while each tweet will contain less than 1\% of these tokens. Therefore, the sparse matrix stored in a compressed form in which just indices and values of non-zero elements are stored as:
\[ M(r,c) = value \]
where $r$ and $c$ show the index of the element where $value \neq 0.$
\begin{table}[H]
	\centering
	\caption{\label{tab:Features_Combinations}Twenty-Four Unique Combinations of Features Used for Training Classifiers}
	\begin{tabularx}{360pt}{llll} \hline 
		S\# & Measurement & N-grams Features & Rule Based Features Appended \\ \hline 
		1 & tf & Uni-grams &  No \\ 
		2 & tf & Bi-grams & No \\  
		3 & tf & Tri-grams  & No \\  
		4 & tf & Uni and Bi-grams  & No \\ 
		5 & tf & Bi and Tri-grams  & No \\
		6 & tf &Uni, Bi and Tri-grams  & No \\  
		7 & tf & Uni-grams & Yes \\ 
		8 & tf & Bi-grams  & Yes \\  
		9 & tf & Tri-grams  & Yes \\  
		10 & tf & Uni and Bi-grams  & Yes \\  
		11 & tf & Bi and Tri-grams  & Yes \\
		12 & tf & Uni, Bi and Tri-grams & Yes \\  
		13 & tf-idf & Uni-grams & No \\ 
		14 & tf-idf & Bi-grams  & No \\  
		15 & tf-idf & Tri-grams & No \\  
		16 & tf-idf & Uni and Bi-grams  & No \\  
		27 & tf-idf & Bi and Tri-grams  & No \\
		18 & tf-idf & Uni, Bi and Tri-grams  & No \\  
		19 & tf-idf & Uni-grams & Yes \\ 
		20 & tf-idf & Bi-grams &  Yes \\  
		21 & tf-idf & Tri-grams &  Yes \\  
		22 & tf-idf & Uni and Bi-grams  & Yes \\ 
		23 & tf-idf & Bi and Tri-grams  & Yes \\ 
		24 & tf-idf & Uni, Bi and Tri-grams  & Yes \\ \hline 
	\end{tabularx}
\end{table}

\begin{algorithm}[H]
	\begin{applyredcolor}
		\DontPrintSemicolon
		\label{algo:rweetseries}
		\KwInput{Set of feature matrices along with labels}
		\KwOutput{Categorized tweets}
		\KwParameters{$TW$: Actual tweets data, $TW_f$: Sets of feature matrices along with labels, $tw_{f}$: Single feature matrix, clf: Machine learning classifier, $y_{id}$: Actual labels for identification of rweets, $\hat{y}_{id}$: Predicted labels for identification of rweets, $y_{ct}$: Actual labels for categorization of rweets, $\hat{y}_{ct}$: Predicted labels for categorization of rweets, $indices$: Indices for non rweets, $TW_{rw}$: Filtered rweets,  $TW_{ct}$: Categorized rweets     }

		$TW_f[R,C] \leftarrow$ Load($TW_f[R,C]$) \; 
		\label{algo:rweetseries:Load} \tcp*{For $TW_f[R,C]$ see line 21 in algorithm 2}
		$clf \in \{SVM, GB, RF, DT, MLP, LR, NB\}$ \; \tcp*{ Acronyms are defined in Table \ref{tab:Classifcation_Models}}
		
		\ForEach{$k,tw_{f}[r, c] \in TW_f[R,C]$}
		{	
			\tcc{Rweet Identification}
			TrainingClassifier(clf, $tw_{f}[r, c]$, $y_{id}[r, d]$) \; \label{algo:rweetseries:identficationTraining}
			$\hat{y}_{id}[r, d] \leftarrow clf.predict(tw_{f}[r, c])$\;  \label{algo:rweetseries:identification} \tcp*{d = 1}
			
			\tcc{Rweet Filtering}
			$tw_{f}[\acute{r}, c], indices \leftarrow$  \texttt{FilterRweets($tw_{f}[r, c],  \hat{y}_{id}[r, d]$)}\;
			\label{algo:rweetseries:FilterRweets} 
			$TW_{rw} \leftarrow TW.delete(indices)$ \; 
			\tcp*{Delete list of rows at $indices$ from $TW$}
			$\hat{y}_{id}[\acute{r}, d] \leftarrow \hat{y}_{id}[r,d].delete(indices)$ \;
			$TW_{rw} \leftarrow$ Merge ($TW_{rw}, \hat{y}_{id}[\acute{r},d])$ \;
			
			\tcc{Rweet Categorization}
			$y_{ct}[\acute{r}, d] \leftarrow y_{ct}[r,d].delete(indices)$ \;
			TrainingClassifier(clf, $tw_{f}[\acute{r}, c]$, $y_{ct}[\acute{r},d]$) \; \label{algo:rweetseries:categorizationTraining}
			$\hat{y}_{ct}[\acute{r},d] \leftarrow clf.predict(tw_{f}[\acute{r}, c])$\;   \label{algo:rweetseries:categorization}
			$TW_{ct} \leftarrow$ Merge ($TW_{rw}, \hat{y}_{ct}[\acute{r}, d])$ \; \tcp*{$|TW_{ct}|  \leq  |TW|$}
		}

		\SetKwFunction{FFilterRweets}{FilterRweets}
		\SetKwProg{Fn}{}{:}{}
		\Fn{\FFilterRweets{$tw_{f}[r, c], \hat{y}_{id}[r, d]$}}{ \label{algo:rweetseries:FFilterRweets}
			$indices \leftarrow \{r: \hat{y}_{id}[r, d] = 0, r \in {Z}^+$, 0 represents non rweets \}\;
			$tw_{f}[\acute{r}, c] \leftarrow $ $tw_{f}[r, c].delete(indices)$, \textrm{ where } $\acute{r} \leq r$ \;  
			\tcp*{Delete list of rows at $indices$ from a feature matrix $tw_{f}[r, c]$}
			\KwRet 	$tw_{f}[\acute{r}, c], indices$\;
		}
		\caption{Rweets Identification and Categorization Series}
	\end{applyredcolor}
\end{algorithm}

\begin{table}[H]
	\centering
	\caption{\label{tab:RE_CharExp}Explanation of Special Characters used in Regular Expressions for Rweet Patterns}
	\begin{tabularx}{340pt}{llX} \hline   
		S\# & Special Character & Explanation \\ \hline 
		1 & {\textbackslash}b & Represents the word boundary \\ 
		2 & . & Indicates any character except a newline \\ 
		3 & * & Represents zero or more occurrence of the preceding \\ 
		4 & ? & Represents zero or one occurrence of the preceding \\ 
		5 & {\textbar} & Refers to OR operator \\ 
		6 & {\textbackslash}s & Used to match whitespace characters, i.e., tab, return, etc. \\ 
		7 & {\textbackslash}w & Refers to a word character, i.e., a-z, A-Z, \_ and 0-9 \\ \hline 
	\end{tabularx}
\end{table}
\subsection{Classification}
To tackle the classification problem, we propose a two-phase classification approach as summarizes in algorithm \ref{algo:rweetseries}. First, we aim to develop a classifier responsible to identify rweets (line \ref{algo:rweetseries:identficationTraining} in algorithm \ref{algo:rweetseries}). These identified rweets are then filtered(line \ref{algo:rweetseries:FilterRweets} in algorithm \ref{algo:rweetseries}) and directed towards the second phase. The second classifier is developed that aims to determine the specific type of request mentioned in a rweet (line \ref{algo:rweetseries:categorizationTraining} in algorithm \ref{algo:rweetseries}). As shown in the Figure \ref{fig:classificationseries}, rweets are first identified in rweet identification(line \ref{algo:rweetseries:identification} in algorithm \ref{algo:rweetseries}), and then they are categorized into six types of requests (i.e., \qq{money}, \qq{volunteer}, \qq{cloth}, \qq{shelter}, \qq{medical}, and, \qq{food})(line \ref{algo:rweetseries:categorization} in algorithm \ref{algo:rweetseries}).

The traditional rule-based approach proved to provide better results in detection and identification of similar problems, e.g., question identification, and extraction \cite{cong2008finding,wang2010exploiting}. Therefore, both traditional rule-based as well as sophisticated machine learning approaches have been used for the rweet identification. On other hand, machine learning approach has only been used for rweets categorization.
In a rule-based approach, classification has been performed using regular expression shown in Table  \ref{tab:Rweet_patterns}. These regular expressions contain both tokens (i.e., am, are, bringing etc.) and special characters (i.e, {\textbar}, * etc.). Special characters have different meaning and interpretation\cite{regExp} that are explained in Table \ref*{tab:RE_CharExp}.

\noindent Each rule is a form 

$R_i:P_{i\ }\ =>C$, where

i = $\{$1, 2, 3, . . . ., 18$\}$,

P${}_{i }$ refers to the 18 patterns respectively,

R${}_{i}$ refers to the corresponding rule for P${}_{i}$,

C = $\{$Rweet$\}$ represent the class label that the pattern classify

These rules are checked against each tweet to classify it as a \qq{rweet} and \qq{not\_rweets}. If a tweet satisfies at least one rule then it is considered a rweet. The sequence of tokens in a regular expression is also important because if a tweet contains all the tokens of a regular expression with a different sequence then it would not be considered as rweet. The method shown in algorithm \ref{algo:algoRweetFilter}, presents the pseudo-code for complete procedure behind the rule-based system.

The machine learning approach has also been used for classification problems on Twitter \cite{ozger2014question,li2012question}. Table \ref{tab:Classifcation_Models} shows the list of 7 different classifiers that have been used for rweet identification and categorization.  Stratified k-fold cross-validation (with k=5) was adopted for developing and comparing classification models because it is an effective scheme to reduce both under-fitting and over-fitting problems\cite{kohavi1995study,a2019evaluation}. Stratified k-fold cross-validation is effective and superior to both train-test split and simple k-fold cross-validation techniques because it partitioned the data in such a way so that each partition or fold contains a similar portion of data for each class label. Data was also randomly shuffled during performing the stratified k-fold cross-validation technique\cite{a2019evaluation}.

\begin{algorithm}[H]
	\DontPrintSemicolon
	\label{algo:algoRweetFilter}
	\KwInput{Actual tweets data obtained from Twitter}
	\KwOutput{Sets of rweets  and not rweets}
	\KwParameters{$TW$: Actual tweets data, $RW$: set of rweets, $NRW$: set of not rweets, $SPs$: Set of sequential patterns  }
	$SPs \leftarrow$ set of all sequential patterns\;
	$RW \leftarrow  \emptyset$\;
	$NRW \leftarrow  \emptyset$\;
		\ForEach{tweet tw in TW}
	{
		\For{$\forall \textrm{pattern} \textit{ p} \in SPs$}
		{
			\If{tw satisfies at least one p}
		{
			$RW.add(tw)$\;
		}
		\Else
		{
			$NRW.add(tw)$\;
		}
		
		}
	}
	\caption{Rweet Filter}
\end{algorithm}

\begin{table}[H]
	\centering
	\caption{\label{tab:Classifcation_Models}List of Classifiers}
	\begin{tabularx}{340pt}{lX} \hline 
		S\# & Classifier \\ \hline 
		1 & Decision Tree (DT) \\ 
		2 & Gradient Boosting (GB) \\  
		3 & Logistic Regression (LR) \\  
		4 & Multilayer Perceptron (MLP) \\  
		5 & Na\"{i}ve Bayes (NB) \\ 
		6 & Support Vector Machine (SVM) \\  
		7 & Random Forest (RF) \\ \hline 
	\end{tabularx}
\end{table}

\section{Experiments and Evaluation}\label{sec:experimentsAndEvaluation}
Numerous experiments have been performed for each classifier (i.e., shown in Table \ref{tab:Classifcation_Models}) on each features set of 24 feature combinations(i.e., shown in Table \ref{tab:Features_Combinations}) in a dataset. There are two datasets, i.e., dataset-1 contains 2940 tweets and dataset-2 contains 2707 tweets. As a result, 336(7x24x2) experiments for the machine learning approach, and one for the rule-based approach only on dataset-1 have been performed as a whole. For the sake of conciseness and page restrictions, only the best results for each classifier is reported in this paper. The complete sets of results could be seen at \cite{studyRes}. 
To evaluate the performance deeply and thoroughly, the proposed architecture is evaluated by both micro and macro evaluation metrics. Results are, then, compared with baselines in order to show the effectiveness of the proposed architecture. Unknown results are replaced with \qq{?} while results less than 50\% are replaced with \qq{-}. In the remaining subsections, evaluation matrices and evaluation results are presented.
\subsection{ Evaluation Metrics}
The commonly used evaluation matrices, i.e., precision, recall, F1-measure, and accuracy have been used in this research \cite{buttcher2010information}. The following types of evaluation measures \cite{sokolova2009systematic} have been used in this research to explore the results obtained by RweetMiner in detail.
\par \emph{ Accuracy:} Accuracy is the ratio of the number of correctly predicted samples to the total number of samples and is one of the widely used evaluation metrics. It is calculated as
\[A=\sum_{l\in L}{\frac{\left|{TP}_l\right|+\left|{TN}_l\right|}{\left|{TP}_l\right|+\left|{TN}_l\right|+\left|{FP}_l\right|+\left|{FN}_l\right|}}\] 
\par \emph{Micro-average Measures:} Micro average is a summary measure over all the tweets (document) without regard to class (category) \cite{buttcher2010information}. It gives equal weight to each per-tweet class prediction \cite{manning2008intro}. As micro average is calculated over occurrences, therefore classes with many occurrences are given more importance. It is an effective measure on data with large classes because it treats each and every tweet equally \cite{sokolova2009systematic,manning2008intro}.

\textbf{Precision${}_{micro}$: }Precision${}_{micro}$ is calculated as
\begin{equation} \label{eq:PreMiicro}
P_{\mu}=\frac{\sum_{l\epsilon L}{\left|{TP}_l\right|}}{\sum_{l\epsilon L}{(\left|{TP}_l\right|+\left|{FP}_l\right|)}}
\end{equation}
\textbf{Recall${}_{micro}$: }Recall${}_{micro}$ is calculated as
\begin{equation} \label{eq:RecMiicro}
R_{\mu}=\frac{\sum_{l\epsilon L}{\left|{TP}_l\right|}}{\sum_{l\epsilon L}{(\left|{TP}_l\right|+\left|{FN}_l\right|)}}
\end{equation}
\textbf{F${}_{1}$-Measure${}_{micro}$:} F${}_{1}$-Measure${}_{micro}$ is calculated as
\begin{equation} \label{eq:F1Miicro}
{F1}_{\mu}=2\times \frac{P_{\mu}\times R_{\mu}}{P_{\mu}+R_{\mu}}
\end{equation}

\par \emph{Macro-average Measures:} 
It is the average of summary measures computed for each class. It is calculated by finding different measures for all predictions in each class, and then computing their average normalized by unweighted mean and, thus considered as per-class average \cite{yang1999evaluation}. It treats each class equally by giving them equal weight and is an effective measure in data with small classes  \cite{sokolova2009systematic,manning2008intro}.  

\textbf{Precision${}_{macro}$: }Precision${}_{macro}$ is calculated as

\begin{equation} \label{eq:PreMaacro}
P_M=\frac{1}{\left|L\right|}\times \sum_{l\epsilon L}\frac{{\left|{TP}_l\right|}}{{\left|{TP}_l\right|+\left|{FP}_l\right|}}
\end{equation}

\textbf{Recall${}_{macro}$: }Recall${}_{macro}$ can be calculated as

\begin{equation} \label{eq:RecMaacro}
R_M=\frac{1}{\left|L\right|}\times \sum_{l\epsilon L}\frac{{\left|{TP}_l\right|}}{{\left|{TP}_l\right|+\left|{FN}_l\right|}}
\end{equation}

\textbf{F${}_{1}$-Measure${}_{macro}$: }F${}_{1}$-Measure${}_{macro}$ is calculated as
\begin{equation} \label{eq:F1Maacro}
{F1}_M=2\times \frac{P_M\times R_M}{P_M+R_M}
\end{equation}

\begin{table}
	\centering
	\caption{\label{tab:results_bin1Bin2_RuleBased}Results for Dataset-1 Achieved using Rule-based Approach}
	\begin{tabularx}{240pt}{lllll} \hline  
		Dataset & Accuracy & Precision & Recall & F1-measure \\ \hline
		Dataset-1 & - & \textbf{99.70} & - & 62.69 \\ \hline
	\end{tabularx}
\end{table}
\subsection{Rweet Identification}
Table \ref{tab:results_bin1Bin2_RuleBased} shows results for dataset-1 for detecting rweets using rule-based approach. It shows that sequential patterns used as rule-based features produced satisfactory performance. Specifically, it performed very well in achieving high precision of 99.7\% because there are very few false positives. The recall is very low because of a large number of false negatives. There are two reasons for getting a too high number of false negatives: 1) eighteen sequential patterns (shown in Table \ref{tab:Rweet_patterns}) are insufficient to detect all types of flexible, and unstructured rweets and, 2) absence of sequential patterns for identifying not rweets.

Results presented in Table \ref{tab:compare_classifier_results_req_identifi}, are obtained using machine learning approach. It shows that RweetMiner achieved stable performance for different types of evaluation metrics for various classifiers that have been tested in this research.
Table \ref{tab:results_rweetIdent_Comp} shows the comparison of results obtained by RweetMiner with the results of the baseline papers, i.e., \cite{purohit2013emergency}, and \cite{nazer2016finding}. It demonstrates that an improved F1-measure of 82.38\% is achieved for rweet identification. As compared to the \cite{purohit2013emergency}, the recall and F1-measure are improved significantly by 52.68\% and  36.81\%, respectively, at the cost of a slight drop in precision. These improved results were achieved by training logistic regression with uni-grams+bi-grams+rule-based features. There is also a list of 19 more improved results of F1-measure achieved using different classifiers and features that vary between 80.64\%-82.38\% for rweet identification and could be accessed at \cite{studyRes}. 

\begin{table}[H]
	\setlength\tabcolsep{4pt}
	\centering
	\scriptsize,
	\caption{\label{tab:compare_classifier_results_req_identifi} Comparison of Micro-average and Macro-average Measures' Results for Rweet Identification Achieved using Various Classifiers along with their Corresponding Features}
	\begin{tabularx}{\textwidth}{lXlllllll}
		\hline \rule{0pt}{9pt}   
		CLF &  Feature Set &  
		$P_{\mu}$ &  $P_{M}$  & 
		$R_{\mu}$ &  $R_{M}$  &  
		$F_{\mu}$ &  $F_{M}$  &  
		A \\ \hline
		\rule{0pt}{7pt} DT & TF(Uni+Bi-grams)+Rules & 76.66 & 76.53 & 76.66 & 76.35 & 76.66 & 76.42 & 76.66  \\ 
		GB & TF(Uni+Bi-grams)+Rules & 79.43 & 79.4 & 79.43 & 79.06 & 79.43 & 79.18 & 79.43 \\ 
		LR & TF(Uni+Bi-grams)+Rules &  \textbf{82.38} & 82.33 &  \textbf{82.38} & 82.11 &  \textbf{82.38} & 82.2 &  \textbf{82.38}\\  
		MLP & TF(Uni+Bi+Tri-grams)+Rules & 81.03 & 80.91 & 81.03 & 80.86 & 81.03 & 80.88 & 81.03\\ 
		NB & TF(Uni+Bi+Tri-grams)+Rules & 81.35  & 82.28 & 81.35 & 80.53 & 81.35   & 80.82 & 81.35\\  
		RF & TF(Uni-grams)+Rules & 78.01 & 77.87 & 78.01 & 77.8 & 78.01 & 77.83 & 78.01\\ 
		SVM & \_ & 54.21 & \_ & 54.21 & 50 & 54.21  & - & 54.21\\ \hline 
	\end{tabularx}
\end{table}

\begin{table}[H]
	\centering
	\scriptsize
	\caption{\label{tab:results_rweetIdent_Comp}Comparison of the Rweet Identification Results with Baselines}
	\begin{tabularx}{340pt}{lllll} \hline   
		Method & Accuracy & Precision & Recall & F1-measure \\ \hline 
		RweetMiner & \textbf{82.38} & \textbf{82.38} & \textbf{82.38} & \textbf{82.38} \\ 
		\cite{purohit2013emergency} & ? & 97.9 & 29.7 & 45.57 \\ 
		\cite{nazer2016finding} & 80.28 & 80.28 & 80.28 & 80.28 \\ \hline 
	\end{tabularx}
\end{table}

\subsection{Rweet Categorization}
Table \ref{tab:compare_classifier_results_req_categori} shows the results achieved for different types of evaluation metrics for various classifiers tested along with the corresponding features. It shows that the RweetMiner provided a very stable performance by not affected by large classes, small classes, and imbalanced classes in dataset-2. Table \ref{tab:results_rweetCateg_Comp} shows the comparison of results obtained by RweetMiner with the results of the baseline paper, i.e., \cite{purohit2013emergency}. It outperformed the baseline\cite{purohit2013emergency} by obtaining high values for each evaluation metric, i.e., accuracy, precision, recall, and F1-measures.

An improved F1-measure of 94.95\% was achieved for rweet categorization using logistic regression classifier, and uni-grams+ bi-grams features. Like rweet identification, there is a list consists of 32 more improved results of F1-measure achieved using different classifiers and features that vary between 92.99\%-94.95\% for rweet categorization and could be accessed at \cite{studyRes}.

\begin{table}
	\setlength\tabcolsep{4pt}
	\centering
	\scriptsize,
	\caption{\label{tab:compare_classifier_results_req_categori}Comparison of Micro-average and Macro-average Measures' Results for Rweet Categorization Achieved using Various Classifiers along with their Corresponding Features}
	\begin{tabularx}{\textwidth}{lXlllllll} 
		\hline 
		CLF &  Feature Set & 
		$P_{\mu}$ &  $P_{M}$  &   
		$R_{\mu}$ &  $R_{M}$  &  $F_{\mu}$ &   $F_{M}$  &   A \\ \hline 
		\rule{0pt}{7pt} DT & TF(Uni+Bi+Tri-grams) & 94.18 & 89.25 & 94.18 & 88.89 & 94.18 & 89.06 & 94.18 \\ 
		GB & TF(Uni+Bi-grams)+Rules & 94.87 & 90.82 & 94.87 & 90.28 & 94.87 & 90.48 & 94.87 \\ 
		LR & TF(Uni+Bi-grams) & \textbf{94.95} & \textbf{92.68} & \textbf{94.95} & \textbf{87.66} & \textbf{94.95} & \textbf{90.01} & \textbf{94.95}\\ 
		MLP & TF(Uni+Bi+Tri-grams) & 93.64 & 92.2 & 93.64 & 83.25 & 93.64 & 87.29 & 93.64\\ 
		NB & TF(Uni+Bi-grams)+Rules & 90.17 & 89.22 & 90.17 & 74.97 & 90.17 & 80.94 & 90.17\\
		RF & TF(Uni-grams) & 92.56 & 89.5 & 92.56 & 80.77 & 92.56 & 84.78 & 92.56 \\ 
		SVM & \_ & 69.94 & - & 69.94 & - &  69.94 & - & 69.94 \\ \hline 
		
	\end{tabularx}
\end{table}

\begin{table}
	\centering
	\scriptsize
	\caption{\label{tab:results_rweetCateg_Comp}Comparison of the Rweet Categorization Results with a Baseline}
	\begin{tabularx}{240pt}{lllll} \hline  
		Method & Accuracy & Precision & Recall & F1-measure \\ \hline 
		RweetMiner &  \textbf{94.95} & \textbf{94.95} & \textbf{94.95} & \textbf{94.95} \\ 
		\cite{purohit2013emergency} & ? & 92.8 & 92.9 & 92.84 \\ \hline 
	\end{tabularx}
\end{table}

\section{Conclusion and Future Work}\label{sec:conclusion}
Identifying and classifying tweets appropriately is difficult because they contain noise and stinky data. Moreover, users do not formally post their requests through tweets. First of all, this research defines a request tweet as \qq{rweet} in the context of social networking sites. Three primary types (i.e., declarative, interrogative, and imperative) and two sub-types (i.e., command, and request for imperative) are then defined for these rweets. %Exiting systems neither plan for preprocessing of tweets nor grasp the context of tweets effectively. Subsequently the focus of the proposed approach is also on these two things.
To purify well the data, this research proposes an effective data preprocessing strategy that comprises nine different types of cleaning operations along with their efficient and effective execution order.  %Operations in the data preprocessing is performed in a sequence, therefore they have a strong influence on each other.
%The execution order is very important and greatly affects the process of purifying the data as well as reduces the dimensionality of feature vector space. %It not only affects the data cleansing process but also influence the time complexity of data preprocessing and memory requirements for the preprocessed data.
The execution order is very important because it not only affects the data cleansing process and memory requirements for the preprocessed data but also influences the processing time of data preprocessing. Feature sets of n-grams have been expanded by combining uni-grams, bi-grams, and/or tri-grams with each other and with the rule-based features(generated utilizing sequential patterns) for extracting deep features to widen distances among look-like tweets. This technique greatly helps in interpreting the context of tweets. 
%\par \textcolor{red}{The proposed architecture of RweetMiner suggests storing intermediate data to speed up the development process of multiple machine learning classifiers simultaneously. As the machine learning life cycle is an iterative process, therefore storing preprocessed data after performing the data preprocessing and storing the feature matrices in memory and processing efficient manner after the feature generation process greatly helps in the early development of the machine learning classifiers by eliminating repetitive re-computations of the computationally expensive processes of data processing and feature generation.}
\par As the machine learning life cycle is an iterative process, therefore the proposed architecture suggests storing the preprocessed data and feature matrices for the early development of multiple machine learning classifiers by eliminating repetitive re-computations of the computationally expensive processes of data processing and feature generation.
\par The obtained results provided sufficient improvement over baselines and make a strong reference point for future work. There are many ways in which this work can be extended. Syntax features like the length of tweets in characters/word, prefix span algorithm can be used for mining frequent patterns from rweets and non-rweets. Feature selection methods, i.e., principal component analysis (PCA), chi-square test, and information gain can be used to reduce the curse of dimensionality, increasing the generalization of classifiers, and decreasing training time for classifiers.
\par In the future, we will also develop distributed RweetMiner system in a big data situation and explore its scalability. Instead of textual contents, tweets also contain other types of data, i.e., videos and images. This data can play an important role in the disastrous situation, therefore it will also be considered. We will also enhance the existing system by utilizing deep learning language models, e.g., BERT, GPT2, and XLNet, etc.

\bibliography{references-list}

\end{document}